\documentclass[runningheads]{llncs}

\usepackage{eccv}

\usepackage{eccvabbrv}

\usepackage{graphicx}
\usepackage{booktabs}
\usepackage{wrapfig}
\usepackage{pifont}
\usepackage{array}
\newcommand{\xmark}{\ding{55}}
\newcommand{\cmark}{\ding{51}}

\usepackage[accsupp]{axessibility}  % Improves PDF readability for those with disabilities.

\usepackage{orcidlink}

\begin{document}

\title{ESCAPE: Episodic Spatial Memory and Adaptive Execution Policy for Long-Horizon Mobile Manipulation}

\titlerunning{ESCAPE}

\author{
Jingjing Qian\inst{1} \and 
Zeyuan He\inst{1} \and 
Chen Shi\inst{1} \and 
Lei Xiao\inst{2} \and 
Li Jiang\inst{1\dagger}
}

\authorrunning{J. Qian et al.}

\institute{\textsuperscript{1}The Chinese University of Hong Kong, Shenzhen \quad \textsuperscript{2}LiAuto Inc.}

\maketitle

{
    \renewcommand{\thefootnote}{$\dagger$}
    \footnotetext{Corresponding Author.}
}

\begin{abstract}
  Coordinating navigation and manipulation with robust performance is essential for embodied AI in complex indoor environments. However, as tasks extend over long horizons, existing methods often struggle due to catastrophic forgetting, spatial inconsistency, and rigid execution. To address these issues, we propose \textbf{ESCAPE} (\textbf{E}pisodic \textbf{S}patial memory \textbf{C}oupled with an \textbf{A}daptive \textbf{P}olicy for \textbf{E}xecution), operating through a tightly coupled perception-grounding-execution workflow. For robust perception, ESCAPE features a Spatio-Temporal Fusion Mapping module to autoregressively construct a depth-free, persistent 3D spatial memory, alongside a Memory-Driven Target Grounding module for precise interaction mask generation. To achieve flexible action, our Adaptive Execution Policy dynamically orchestrates proactive global navigation and reactive local manipulation to seize opportunistic targets. ESCAPE achieves state-of-the-art performance on the ALFRED benchmark, reaching 65.09\% and 60.79\% success rates in test seen and unseen environments with step-by-step instructions. By reducing redundant exploration, our ESCAPE attains substantial improvements in path-length-weighted metrics and maintains robust performance (61.24\%/56.04\%) even without detailed guidance for long-horizon tasks.

\end{abstract}

\section{Introduction} \label{sec:intro}
The vision of autonomous robots capable of assisting humans in everyday household tasks has long been a driving force in robotics research~\cite{levine2018learning, pashevich2021episodic, min2021film, brohan2022rt}. Recent advances in Embodied AI~\cite{prabhudesai2020embodied, stepputtis2020language, song2023llm, chi2025diffusion, kim2024openvla, luthinkbot} have demonstrated remarkable progress in enabling artificial agents to perceive, navigate, and interact with their surroundings. Within this context, embodied Long-Horizon Mobile Manipulation~\cite{shridhar2020alfred, homerobotovmm} emerges as a crucial next step, requiring agents to coordinate both long-term navigation and sophisticated object manipulation while maintaining robust performance in dynamic, complex indoor environments.

Embodied Long-Horizon Mobile Manipulation poses key challenges for autonomous agents, particularly as tasks shift from isolated actions to complex, \textit{long-horizon} goals. These tasks demand robust spatial reasoning to understand complex and dynamic 3D environments from limited egocentric views, as well as sophisticated planning to coordinate long-term navigation with precise object manipulation while maintaining goal-oriented behavior. Real-time adaptation to environmental feedback further adds to the complexity. ALFRED~\cite{shridhar2020alfred} benchmark serves as a standard testbed for evaluating these challenges, requiring agents to follow natural language instructions to complete long-horizon household tasks involving extensive exploration and opportunistic interaction.

\begin{figure}[t]
  \centering
  \includegraphics[width=0.99\textwidth]{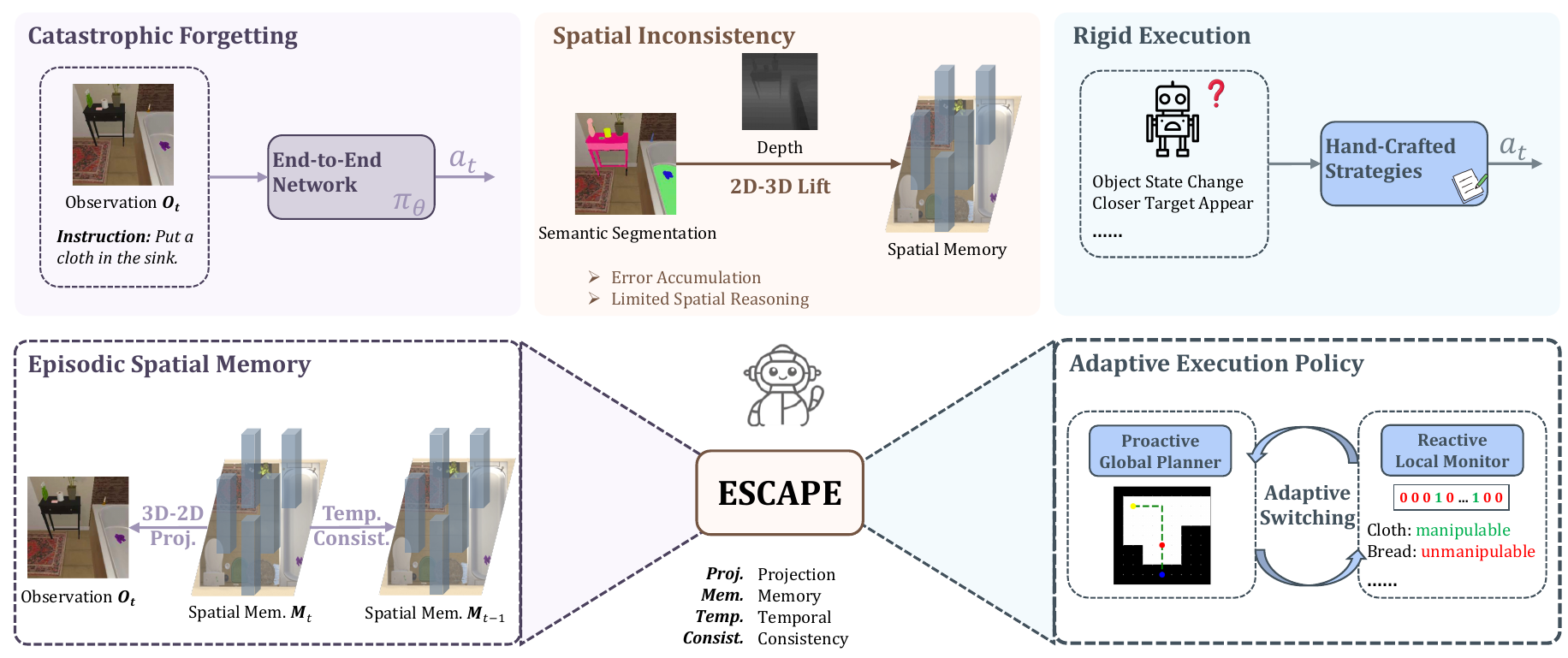}
  \caption{Existing methods struggle with long-horizon tasks due to catastrophic forgetting, spatial inconsistency, and rigid execution. ESCAPE addresses these issues through a persistent episodic spatial memory that captures spatio-temporal relationships, coupled with an adaptive execution policy that seamlessly coordinates proactive global navigation and reactive local manipulation.}
  \label{fig:intro}
  \vspace{-10pt}
\end{figure}

Despite recent progress, executing long-horizon mobile manipulation tasks remains exceptionally challenging, exposing three critical limitations in existing methods, as illustrated in Fig. \ref{fig:intro}. First, \textbf{catastrophic forgetting} occurs in end-to-end approaches (e.g., Seq2Seq~\cite{shridhar2020alfred}, MOCA~\cite{singh2021factorizing}, and E.T.~\cite{pashevich2021episodic}) that directly map current observations to actions without maintaining a persistent spatial memory. This forces agents to redundantly re-explore previously visited areas when searching for target objects, drastically reducing long-horizon efficiency.

Second, \textbf{spatial inconsistency} plagues modular methods like FILM~\cite{min2021film}, CAPEAM~\cite{kim2023context}, DISCO~\cite{xu2024disco}, and ThinkBot~\cite{luthinkbot}. These methods process semantics and affordances independently in 2D image before back-projecting them into 3D space using imperfect depth estimation. In long-horizon tasks, these perception errors gradually accumulate due to inaccurate depth-semantic lifting. Furthermore, restricting feature interactions entirely to the 2D image plane fundamentally limits the agent's 3D spatial reasoning capabilities. 

Third, \textbf{rigid execution} limits the generalization of methods (e.g., FILM~\cite{min2021film}, Prompter~\cite{inoue2022prompter}, and CAPEAM~\cite{kim2023context}) that rely heavily on hand-crafted strategies for specific scenarios, such as object state changes and closer target appearances. By blindly following fixed long-term trajectories, these methods fail to react promptly and cannot preemptively interrupt navigation when encountering a target earlier than expected, leading to suboptimal execution paths.

To address these key issues, we propose \textbf{ESCAPE}, a memory-centric framework designed for long-horizon mobile manipulation, as shown in Fig. \ref{fig:intro}. ESCAPE operates through a tightly coupled perception $\rightarrow$ grounding $\rightarrow$ execution workflow. In the \textbf{perception} phase, our \textit{Spatio-Temporal Fusion Mapping} module leverages spatio-temporal cross-attention operating directly in 3D space to fuse current observation with historical memory, autoregressively constructing a persistent episodic spatial memory that captures temporal consistency and spatial-semantic relationships. Notably, by using precise 3D-to-2D projection to extract visual features, this module eliminates depth estimation dependence, enhancing the agent's 3D spatial reasoning capabilities. For the subsequent \textbf{grounding} phase, a \textit{Memory-Driven Target Grounding} module uses the 3D geometric features stored in the memory to query current 2D visual observations, extracting precise masks for physical interaction. Finally, in the \textbf{execution} phase, our \textit{Adaptive Execution Policy} resolves the conflict between long-term planning and short-term reacting. It employs a proactive global planner for memory-based navigation, while running a reactive local monitor concurrently to seize immediate manipulation targets, seamlessly shifting the execution horizon as needed. For fair evaluation on ALFRED~\cite{shridhar2020alfred}, we adopt~\cite{min2021film}'s language parsing module while focusing on advancing navigation and manipulation abilities.

In summary, our main contributions are three-fold:
\begin{itemize}
\item We propose a novel Episodic Spatial Memory framework tailored for long-horizon tasks, featuring a Spatio-Temporal Fusion Mapping module for depth-free, consistent 3D spatial memory construction, and a Memory-Driven Target Grounding module for precise interaction mask generation.
\item We introduce an Adaptive Execution Policy that dynamically orchestrates proactive global navigation and reactive local manipulation, improving the operational efficiency and adaptability of the embodied agent.
\item ESCAPE achieves superior performance on the highly competitive ALFRED benchmark. Notably, by reducing redundant exploration, it attains substantial improvements in path-length-weighted metrics (PLWSR and PLWGC), demonstrating highly efficient long-horizon execution.
\end{itemize}

\section{Related Work}
\textbf{Embodied Mobile Manipulation.} Enabling mobile robots to navigate and manipulate objects in complex environments remains a major embodied AI challenge. Various simulators and benchmarks have emerged to advance this field~\cite{kolve2017ai2,batra2020objectnav,srivastava2022behavior,szot2021habitat,ehsani2024spoc}. While early works focused on navigation in static scenes, recent benchmarks require agents to manipulate objects in dynamic environments with evolving semantics. Some support physical robot testing~\cite{homerobotovmm, lambdabenchmark}, while most use simulated environments for reproducible evaluation~\cite{cheng2025embodiedeval,yang2025embodiedbench}. ALFRED~\cite{shridhar2020alfred} stands out by integrating navigation and manipulation, challenging agents to follow natural language instructions for \textit{long-horizon} household tasks. Its dynamic environments, which require extensive exploration and opportunistic interaction, provide an ideal testbed for our proposed ESCAPE framework, which addresses these challenges by coupling a persistent episodic spatial memory with an adaptive execution policy for flexible navigation and manipulation.

\noindent \textbf{Scene Spatial Memory.} Spatial memory is crucial for long-horizon mobile manipulation. Some approaches rely on explicit memory structures like 2D semantic maps or topological graphs~\cite{zhu2021soon, zhang2025hoz++}. While intuitive, these discrete representations struggle to preserve fine-grained 3D geometry and suffer from error accumulation during prolonged navigation. Conversely, implicit representations like Neural Radiance Fields (NeRF)~\cite{mildenhall2021nerf, liu2023semantic, zhu2023i2} and 3D Gaussian Splatting (3DGS)~\cite{kerbl3Dgaussians, yu2024mip, chen2024survey} enable high-fidelity 3D reconstruction and rendering. However, their intensive optimization and implicit formulation complicate real-time semantic querying for physical interaction. To bridge this gap, Bird's-Eye-View (BEV) representations~\cite{huang2021bevdet, can2021structured, philion2020lift, bevfusion,wang2022detr3d, petr, pan2020cross} offer an effective top-down spatial memory paradigm. Utilizing spatio-temporal transformers~\cite{li2024bevformer} for historical feature aggregation, BEV enables rich spatial-semantic fusion. Building on this, ESCAPE constructs a persistent episodic spatial memory, elegantly balancing 3D spatial reasoning with efficient 2D target grounding.

\noindent \textbf{Adaptive Action Planning.} In long-horizon mobile manipulation, agents must robustly translate complex perceptual states into executable actions~\cite{zhao2024epo, wu2024embodied}. FILM~\cite{min2021film} utilizes neural networks to generate actions from semantic maps, though such map-based methods often lack precision for object grounding. To address this, DISCO~\cite{xu2024disco} introduces fine-action networks that improve alignment and reduce ambiguity. However, despite these perceptual and grounding advancements, both methods rely on rigid sequential planning. This inflexibility prevents agents from preemptively interrupting navigation to seize opportunistic manipulation targets. To address this issue, our Adaptive Execution Policy couples a proactive global planner with a reactive local monitor, dynamically shifting between long-horizon navigation and immediate manipulation.

\section{Method} \label{sec:all}
\begin{figure}[t]
  \centering
  \includegraphics[width=0.99\textwidth]{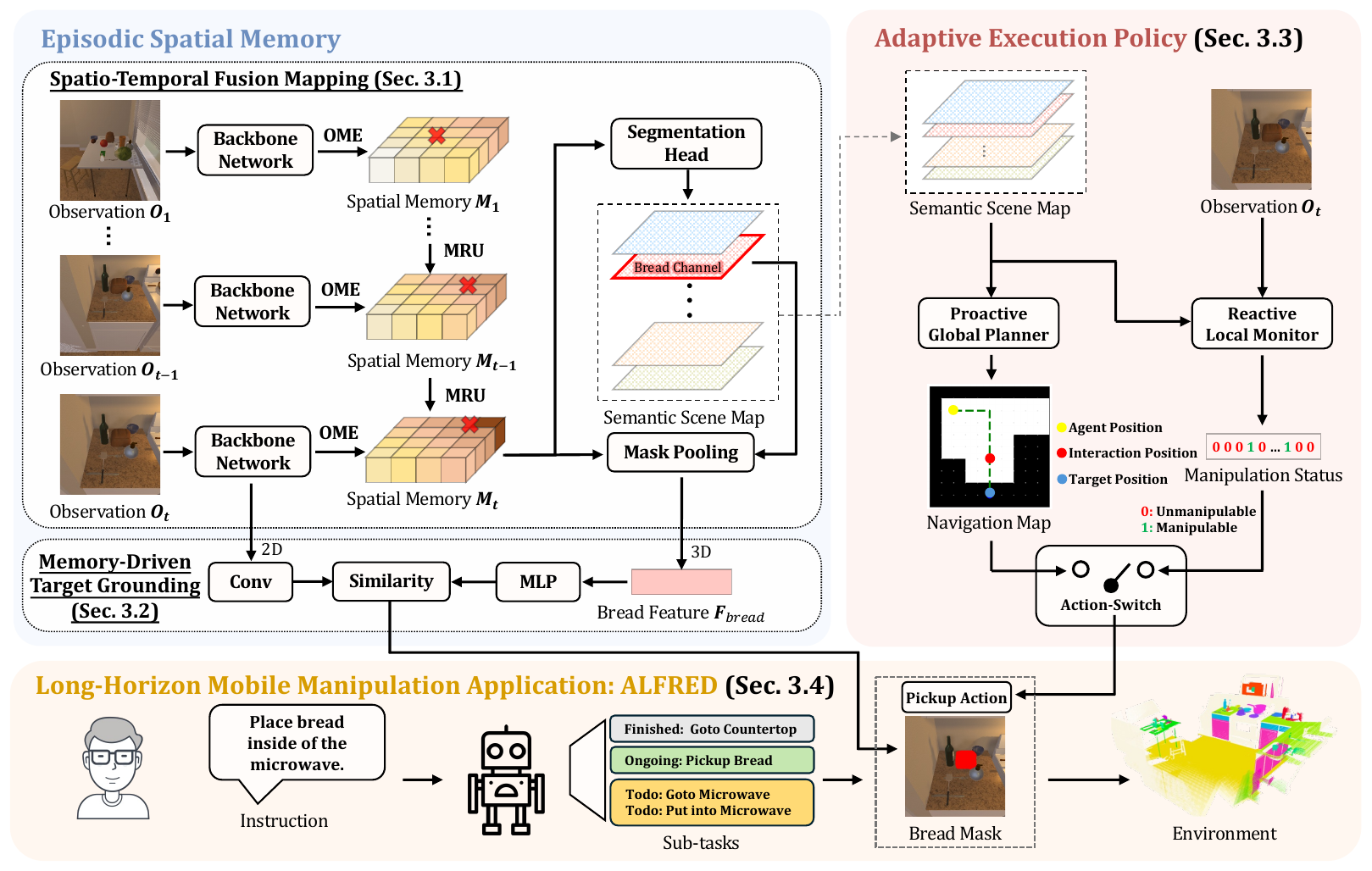}
  \caption{Our ESCAPE framework features two core pillars: a unified Episodic Spatial Memory that processes observations via Spatio-Temporal Fusion Mapping module and Memory-Driven Target Grounding module to generate semantic scene maps and precise interaction masks, and an Adaptive Execution Policy that dynamically shifts between Proactive Global Planner and Reactive Local Monitor. The application shows placing bread inside a microwave, where the agent (marked by \textcolor{red}{\xmark}) navigates and manipulates objects based on instructions.}
  \label{fig:method}
  \vspace{-10pt}
\end{figure}

In this work, we present our proposed ESCAPE. The overall architecture is shown in Fig. \ref{fig:method}. To address the catastrophic forgetting and rigid execution typically encountered in long-horizon embodied tasks, ESCAPE is built upon two core pillars: a unified \textbf{Episodic Spatial Memory} system for robust perception, and an \textbf{Adaptive Execution Policy} for flexible action.

Specifically, this memory-centric framework forms a tightly coupled sequential perception $\rightarrow$ grounding $\rightarrow$ execution workflow. First, the \textit{Spatio-Temporal Fusion Mapping} module (Sec. \ref{sec:mapping}) autoregressively constructs and updates a persistent 3D spatial memory from egocentric observations. Based on this globally consistent memory, the \textit{Memory-Driven Target Grounding} module (Sec. \ref{sec:grounding}) uses 3D object features derived from the memory to query 2D visual observations, extracting precise masks for physical interactions. Finally, the \textit{Adaptive Execution Policy} (Sec. \ref{sec:policy}) leverages the spatial memory for proactive long-horizon navigation trajectories, while a reactive local monitor simultaneously seizes immediate manipulation opportunities. We conclude this section by detailing ESCAPE's application on the ALFRED~\cite{shridhar2020alfred} benchmark (Sec. \ref{sec:alfred}).

\subsection{Spatio-Temporal Fusion Mapping} \label{sec:mapping}
For long-horizon mobile manipulation tasks, an effective memory representation must capture spatial relationships (such as relative object positions) and semantic information, while maintaining persistent temporal consistency across dynamic environments to avoid redundant exploration. Inspired by recent BEV representations~\cite{li2024bevformer,huang2021bevdet} that employ spatio-temporal transformers, we propose the Spatio-Temporal Fusion Mapping module. Unlike original BEV methods using multi-view images, our module autoregressively aggregates temporal images from a single egocentric camera to construct a persistent 3D spatial memory.

To achieve accurate and efficient memory construction, we introduce an Observation-to-Memory Encoding (OME) mechanism implemented via spatial cross attention~\cite{li2024bevformer} to update the memory feature from the current observation. Additionally, to integrate historical memory and acquire spatial reasoning ability in 3D space, we propose a Memory Retrieval and Update (MRU) mechanism utilizing temporal cross attention. Finally, we employ a 3D map semantic segmentation head to convert memory features into a semantic map.

\begin{wrapfigure}{r}{0.47\textwidth}
  \centering
  \vspace{-20pt}
  \includegraphics[width=\linewidth]{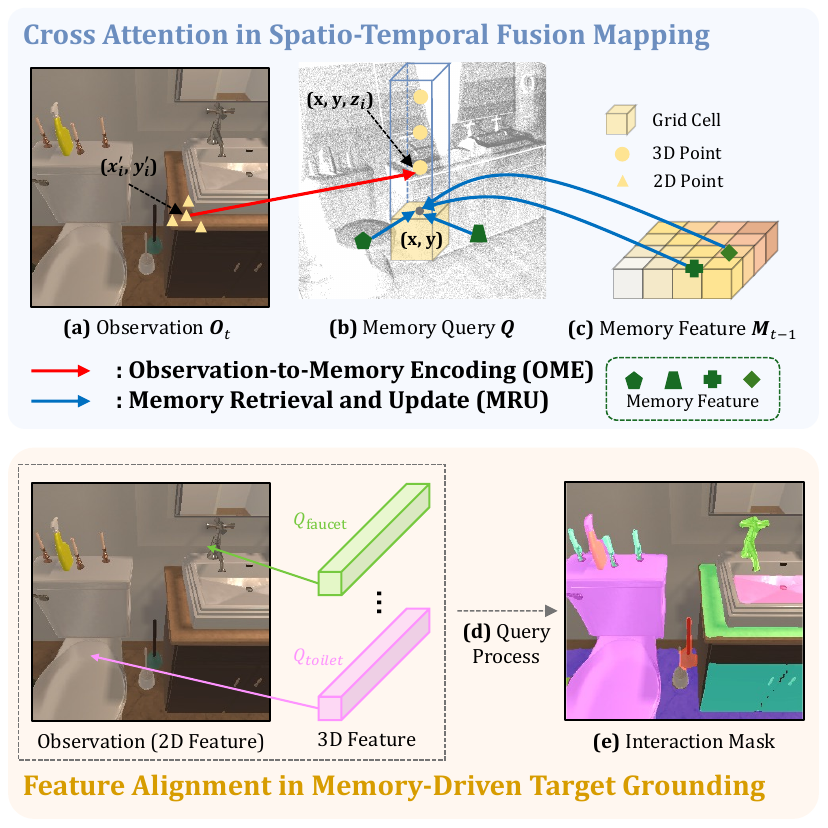}
  \small
  \caption{Demonstration of cross attention in the Spatio-Temporal Fusion Mapping and 2D-3D feature alignment in the Memory-Driven Target Grounding.}
  \vspace{-20pt}
  \label{fig:scene}
\end{wrapfigure}

\noindent \textbf{Observation-to-Memory Encoding (OME).} As shown in Fig. \ref{fig:scene} (a) and (b), from a bird's-eye perspective, we partition the entire room into an $H \times W$ grid, where each grid cell is initialized with a memory query $\textbf{Q}_p \in \mathbb{R}^{1 \times C}$ located at $p = (x,y)$ with $C$ dimension. These queries are then lifted into pillar-like queries, where each 2D grid $p = (x,y)$ is transformed into $\textbf{N}_{\text{ref}}$ 3D reference points $(x,y,z_i)$ sampled along the height dimension. For each reference point, we project it onto the current observation through the projection matrix of the agent's camera, which can be written as:
\begin{equation}
z'_i \cdot \begin{bmatrix} x'_i & y'_i & 1 \end{bmatrix}^T = \textbf{P} \cdot \begin{bmatrix} x & y & z_i & 1 \end{bmatrix}^T.
\end{equation}
Here, $(x'_i, y'_i)$ is the 2D image plane coordinate projected from 3D point $(x, y, z_i)$, $\textbf{P} \in \mathbb{R}^{3 \times 4}$ is the known projection matrix of agent's camera, and $z'_i$ represents the depth factor of the projected point in the camera coordinate system. This projection yields a 2D reference point in the image plane, around which we sample and aggregate image features through weighted summation to obtain the output of OME, formulated as follows:
\begin{equation}
\label{eq:ome}
\text{OME}(\textbf{Q}_p, \textbf{F}_t) = \sum_{i=1}^{\textbf{N}_{\text{ref}}} \text{DeformAttn}(\textbf{Q}_p, (x_i', y_i'), \textbf{F}_t),
\end{equation}
where $i$ indexes the reference points and $\textbf{F}_t$ is the multi-level semantic features extracted from the current frame $\textbf{O}_t$ using a ResNet50 backbone~\cite{he2016deep}. After applying this computation process to all queries in the spatial memory, we finish updating the global memory feature using the current observation.

Here, two points require clarification. First, vanilla multi-head attention~\cite{vaswani2017attention} in our OME is computationally expensive when applied globally across the entire spatial grid. Following~\cite{li2024bevformer}, we adopt deformable attention (DeformAttn in Eq.~\ref{eq:ome})~\cite{zhu2020deformable}, where each memory query $\textbf{Q}_p$ only interacts with its regions of interest in the observation. Second, previous methods like FILM~\cite{min2021film}, DISCO~\cite{xu2024disco} and ThinkBot~\cite{luthinkbot} back-project 2D image features to 3D space using depth estimation, causing error accumulation in long-horizon mobile manipulation tasks. Our OME instead projects 3D points to the image plane to extract 2D features for 3D spatial memory updates, eliminating depth estimation dependence and improving memory fidelity.

\noindent \textbf{Memory Retrieval and Update (MRU).} As shown in Fig. \ref{fig:scene} (b) and (c), given the memory query $\textbf{Q}$ at current timestep $t$ and the cached spatial memory $\textbf{M}_{t-1}$ from timestep $t-1$, each grid query $\textbf{Q}_p$ at position $p = (x, y)$ retrieves relevant features from $\textbf{M}_{t-1}$ to prevent catastrophic forgetting while interacting with neighboring features in $\textbf{Q}$ to enable spatial reasoning. For computational efficiency, we adopt deformable attention to implement MRU, similar to OME, which can be formulated as:
\begin{equation}
\text{MRU}(\textbf{Q}_p, \{\textbf{Q}, \textbf{M}_{t-1}\}) = \sum_{\textbf{V} \in \{\textbf{Q}, \textbf{M}_{t-1}\}} \text{DeformAttn}(\textbf{Q}_p, p, \textbf{V}).
\end{equation}
Our MRU leverages temporal cross attention to adaptively sample and aggregate memory features from the previous timestep, maintaining long-term consistency. Meanwhile, each grid query interacts with surrounding memory features in 3D space, which better captures geometric relationships and enhances spatial reasoning such as relative positions and distances.

\noindent \textbf{3D Map Semantic Segmentation Head.} As shown in Fig. \ref{fig:method}, our Spatio-Temporal Fusion Mapping module processes the current observation via a backbone to generate image features, then uses OME to extract features for memory updates while MRU fuses temporal and 3D spatial information, producing the updated episodic memory $\textbf{M}_t \in \mathbb{R}^{H \times W \times C}$, where $H \times W$ denotes grid cells and $C$ represents feature dimension. Based on this memory feature, we introduce a 3D map semantic segmentation head to predict object category distribution on the grid map $\textbf{D}_t$, trained with ground-truth labels $\textbf{L}_t$ through:
\begin{equation}
\mathcal{L}_{\text{map}} = \frac{1}{HW} \sum_{h=1}^{H} \sum_{w=1}^{W} \text{BCE}(\textbf{D}_t^{h,w}, \textbf{L}_t^{h,w}),
\end{equation}
where $\mathcal{L}_{\text{map}}$ is the map segmentation loss, $\textbf{L}_t$ is a multi-hot vector for each grid cell indicating the object categories present in that cell and BCE denotes binary cross-entropy loss measuring grid-wise discrepancy between predicted probabilities and ground-truth labels per category.

\subsection{Memory-Driven Target Grounding} \label{sec:grounding}
While the Spatio-Temporal Fusion Mapping module effectively constructs a global 3D spatial and semantic memory, it lacks the fine-grained 2D localization required for physical manipulation. In long-horizon tasks, accurate identification and localization of target objects from the agent's egocentric view are crucial. To address this, we introduce the Memory-Driven Target Grounding module, which extracts interaction masks by leveraging the rich, object-specific features already stored within our episodic memory. Inspired by recent works~\cite{peng2023openscene, schult2023mask3d} on 2D-3D feature alignment, we propose generating target object masks by querying the current observation with abstract geometric features from the spatial memory.

As shown in Fig.~\ref{fig:method}, for the \textbf{Pickup Bread} sub-task, we perform mask pooling over the memory feature $\textbf{M}_t$ using the bread channel from the semantic scene map. We aggregate features from bread-containing grid cells via averaging to compute $\textbf{F}_{bread}$, then apply a multi-layer perception (MLP) to generate the bread query $\textbf{Q}_{bread}$. The general formula for computing object queries is:
\begin{equation}
\begin{split}
\textbf{F}_{object} &= \frac{1}{|\mathcal{G}_{object}|} \sum_{(i,j) \in \mathcal{G}_{object}} \textbf{M}_t^{(i,j)}, \\
\textbf{Q}_{object} &= \text{MLP}(\textbf{F}_{object}),
\end{split}
\end{equation}
where $\mathcal{G}_{object}$ denotes grid cells where the object channel equals 1. Since object queries are derived from the episodic memory, they encode global geometric properties and are termed 3D object features.

Simultaneously, convolutional layers are used to map backbone image features to new 2D object features. The similarity computation between 3D object queries and 2D object features produces object masks, precisely localizing objects for interaction. The toilet scene in Fig. \ref{fig:scene} (d) and (e) further shows this 2D-3D feature alignment mechanism. To train the Memory-Driven Target Grounding module, we introduce a 2D image semantic segmentation loss $\mathcal{L}_{\text{img}}$ as follows:
\begin{equation}
\mathcal{L}_{\text{img}} = \frac{1}{HW} \sum_{h=1}^{H} \sum_{w=1}^{W} \text{BCE}(\textbf{P}_t^{h,w}, \textbf{G}_t^{h,w}),
\end{equation}
\noindent where $\textbf{P}_t^{h,w}$ represents the predicted image mask at pixel $(h,w)$ at time-step $t$, and $\textbf{G}_t^{h,w}$ denotes the corresponding ground-truth segmentation label.

\noindent \textbf{Joint Training Objective and Implementation.} To enable the episodic memory to capture spatial relationships while empowering accurate target grounding, we propose joint supervision combining 3D map segmentation and 2D image segmentation losses. The total loss $\mathcal{L}$ is formulated as follows:
\begin{equation}
\mathcal{L} = \mathcal{L}_{\text{map}} + \lambda\mathcal{L}_{\text{img}},
\end{equation}
where $\lambda$ is a hyperparameter that balances the two segmentation losses.

Following DISCO~\cite{xu2024disco}, we collect training instructions, images, semantic map and image labels from ALFRED~\cite{shridhar2020alfred} expert trajectories. For temporal continuity, semantic map labels at timestep $t$ incorporate object distributions from step $1$ to current timestep. We use a ResNet50~\cite{he2016deep} backbone, OME and MRU layers with deformable attention, a map segmentation head and a image segmentation head. On ALFRED, each scene is modeled as a $25m \times 25m$ room with $25cm \times 25cm$ grids, resulting in memory queries of size $H \times W \times C$ where $H = W = 100$ and $C = 256$. These memory queries serve as learnable parameters that are jointly optimized during training. The model is trained using AdamW with BCE loss for 25 epochs, batch size 64 and learning rate 2e-4. Training completes in 24 hours on 8 RTX 4090 GPUs.

\subsection{Adaptive Execution Policy} \label{sec:policy}
\begin{wrapfigure}{l}{0.46\textwidth}
  \centering
  \vspace{-20pt}
  \includegraphics[width=\linewidth]{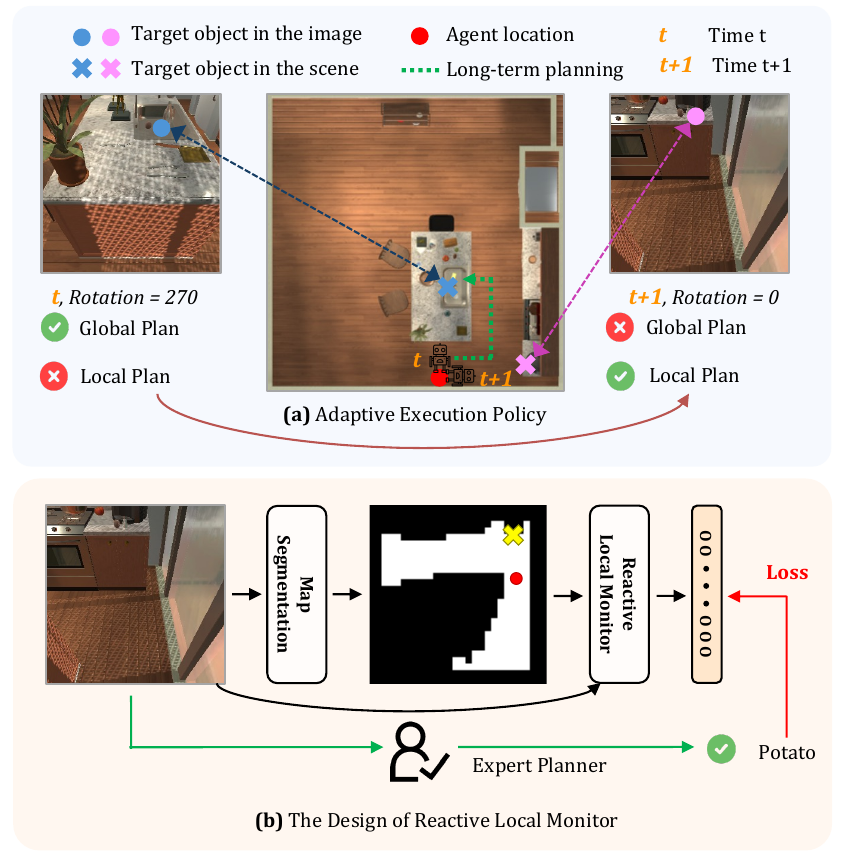}
  \caption{Demonstration of the Adaptive Execution Policy and our design of reactive local monitor.}
  \vspace{-20pt}
  \label{fig:policy}
\end{wrapfigure}

In long-horizon mobile manipulation tasks, there is a fundamental conflict between executing a prolonged navigation plan and reacting to suddenly appearing environmental opportunities. As shown in Fig. \ref{fig:policy} (a), during a \textbf{Pickup Potato} sub-task, after detecting a potato and proactively planning a long route, the agent might discover another potato in a previous blind spot after turning right. Rigidly following the initial plan reduces execution efficiency. To address this, we propose the Adaptive Execution Policy, a dynamic execution mechanism combining a proactive global planner and a reactive local monitor.

\noindent \textbf{Random Exploration.} Before potato detection, the agent explores based on a navigable map. In the semantic map generated by the spatial memory, a grid is considered \textbf{navigable} if only the floor category has a value of 1 while all other object categories are 0, thereby generating the navigable map. The agent randomly selects destinations from navigable cells and plans paths BFS (Breadth-First Search) algorithm.

\noindent \textbf{Proactive Global Planner.} After potato detection, the agent utilizes its persistent spatial memory for proactive goal navigation. It locates the potato at the highest probability grid, expands with navigable grids within a 1-meter radius, and uses BFS to generate long-horizon navigation waypoints (grid coordinates) considering potato region, agent position, and navigable area. Thus, the proactive global planner is a rule-based module requiring no training.

\noindent \textbf{Reactive Local Monitor.} Due to the agent's limited field of view, the global planner may overlook information in occluded areas, causing unnecessary travel. To address this, we introduce a reactive local monitor that continuously screens the short-horizon environment for immediate manipulation opportunities while executing the long-horizon navigation plan. As shown in Fig. \ref{fig:policy} (a), when discovering a potato within range during navigation, the local monitor preempts the global planner, bypassing the remainder of the long-horizon path to execute an immediate pickup action.

As shown in Fig. \ref{fig:policy} (b), our reactive local monitor takes current observations and semantic scene maps as inputs, outputting a binary vector indicating whether each object category is manipulable (1=manipulable, 0=unmanipulable). An object is manipulable only if semantically manipulable and within operational range, requiring precise understanding of semantic information in observation and spatial relationships in the semantic scene map. To train this monitor, we design an expert planner with full scene knowledge for data collection. Following ALFRED's~\cite{shridhar2020alfred} data generation process, our expert planner annotates manipulable target categories per frame, yielding 332,432 training images.

The reactive local monitor employs ResNet50~\cite{he2016deep} as backbone for visual feature and map feature extraction, followed by a linear classifier predicting manipulable object categories. The model is trained using BCE loss based on manipulation labels from the expert planner. Training utilizes the AdamW optimizer with learning rate of $5 \times 10^{-5}$ and batch size of 128 for 40 epochs to achieve optimal performance. Training completes in 2 hours on 8 RTX 4090 GPUs.

Our Adaptive Execution Policy systematically generates actions by leveraging both modules. For multi-instance targets (potatoes, watches, apples), the reactive local monitor opportunistically discovers new goals to improve efficiency. For unique large targets (fridges, sidetables, sofas), the episodic spatial memory provides precise localization, making the proactive global routes highly accurate. When the reactive local monitor fails due to occlusion or inappropriate pose, our hierarchical fallback strategy continues executing existing long-horizon plans or generates new navigation plans to reachable targets.

\subsection{Long-Horizon Mobile Manipulation Application: ALFRED} \label{sec:alfred}
We evaluate our approach on ALFRED~\cite{shridhar2020alfred}, a widely-adopted benchmark for long-horizon embodied instruction following. More introduction about the benchmark can be found in Appendix \ref{sec:benchmark}. Each ALFRED task contains high-level goal descriptions and step-by-step instructions. Appendix \ref{sec:alfred_example} shows an ALFRED task \textbf{Put a cooked potato slice on the counter} where our agent only needs to place any cooked potato slice on any counter to complete the task. This allows our approach to focus on semantic segmentation to identify object categories rather than distinguishing individual instances.

Fig. \ref{fig:method} illustrates the ESCAPE pipeline on ALFRED. Based on received instruction, the agent converts it into executable sub-tasks. For fair comparison with baselines~\cite{min2021film, kim2023context, xu2024disco}, we adopt~\cite{min2021film}'s instruction processing module. For each sub-task, the agent relies on its Episodic Spatial Memory to perceive the environment and generate accurate scene maps and interaction masks. Simultaneously, it employs the Adaptive Execution Policy to seamlessly alternate between proactive long-horizon planning and reactive short-horizon interaction for improved task efficiency. Details are in Appendix \ref{sec:application}.

\section{Experiments} \label{sec:experiments}
\subsection{Dataset and Metrics}
We conduct experiments on ALFRED~\cite{shridhar2020alfred}, a benchmark dataset for vision-language navigation and long-horizon interaction tasks. The dataset comprises 25,726 episodes: 21,023 for training, 1,641 for validation, and 3,062 for testing. Both validation and test sets are divided into seen and unseen environments, with validation containing 820/821 episodes and test containing 1,533/1,529 episodes for seen/unseen scenes. We evaluate our method against competitive baselines using the test split, while using the validation split for analytical studies.

We employ four metrics: Success Rate (\textbf{SR}) measures task completion proportion; Goal Condition (\textbf{GC}) measures the percentage of met goal conditions for partial task success; Path Length Weighted Success Rate (\textbf{PLWSR}) and Goal Condition (\textbf{PLWGC}) incorporate efficiency by comparing paths to expert demonstrations. SR and GC assess task effectiveness, while PLW variants assess efficiency. Higher values indicate better performance.

\begin{table}[t]
\centering
\scriptsize
\caption{Performance comparison on ALFRED benchmark~\cite{shridhar2020alfred}. \cmark / \xmark\ denotes whether step-by-step instructions are used. Bold values indicate the best results for each metric.}
\setlength{\tabcolsep}{4pt}
\renewcommand{\arraystretch}{1.3}
\newcommand{\thickhline}{\noalign{\hrule height 1pt}}
\newcommand{\thinhline}{\noalign{\hrule height 0.5pt}}

\begin{tabular*}{\textwidth}{@{\extracolsep{\fill}}
    p{2.3cm}         % 模型名称列
    >{\centering\footnotesize\arraybackslash}p{0.6cm}
    || 
    >{\centering\arraybackslash}p{0.7cm}    % SR列
    >{\centering\arraybackslash}p{0.7cm}    % GC列
    >{\centering\arraybackslash}p{0.9cm}  % PLWSR列
    >{\centering\arraybackslash}p{0.9cm}  % PLWGC列
    || 
    >{\centering\arraybackslash}p{0.7cm}    % SR列
    >{\centering\arraybackslash}p{0.7cm}    % GC列
    >{\centering\arraybackslash}p{0.9cm}  % PLWSR列
    >{\centering\arraybackslash}p{0.9cm}  % PLWGC列
}
\thickhline
&  & \multicolumn{4}{c||}{\textbf{Test Seen}} & \multicolumn{4}{c}{\textbf{Test Unseen}} \\
& Ins. & SR & GC & PLWSR & PLWGC & SR & GC & PLWSR & PLWGC \\
\thinhline

Seq2Seq~\cite{shridhar2020alfred} & \cmark & 4.00 & 9.40 & 2.00 & 6.30 & 0.40 & 7.00 & 0.10 & 4.30 \\
MOCA~\cite{singh2021factorizing} & \cmark & 26.81 & 33.20 & 19.52 & 26.33 & 7.65 & 15.73 & 4.21 & 11.24 \\
E.T.~\cite{pashevich2021episodic} & \cmark & 38.42 & 45.44 & 27.78 & 34.93 & 8.57 & 18.56 & 4.21 & 11.46 \\
ABP~\cite{kim2021agent} & \cmark & 44.55 & 51.13 & 3.88 & 4.92 & 15.43 & 24.76 & 1.08 & 2.22 \\
FILM~\cite{min2021film} & \cmark & 27.67 & 38.51 & 11.23 & 15.06 & 26.49 & 36.37 & 10.55 & 14.30 \\
LGS-RPA~\cite{murray2022following} & \cmark & 40.05 & 48.66 & 21.28 & 28.97 & 35.41 & 45.24 & 15.68 & 22.76 \\
Prompter~\cite{inoue2022prompter} & \cmark & 51.17 & 60.22 & 25.12 & 30.21 & 45.32 & 56.57 & 20.79 & 25.80 \\
CAPEAM~\cite{kim2023context} & \cmark & 51.79 & 60.50 & 21.60 & 25.88 & 46.11 & 57.33 & 19.45 & 24.06 \\
DISCO~\cite{xu2024disco} & \cmark & 59.50 & 66.10 & 40.60 & 47.40 & 56.50 & 66.80 & 36.50 & 44.50 \\
ThinkBot~\cite{luthinkbot} & \cmark & 62.69 & 71.64 & 32.02 & 37.01 & 57.82 & 67.75 & 26.93 & 30.73 \\
\textbf{ESCAPE (Ours)} & \cmark & \textbf{65.09} & \textbf{72.88} & \textbf{52.42} & \textbf{58.29} & \textbf{60.79} & \textbf{68.60} & \textbf{46.82} & \textbf{52.57} \\

\thinhline

HLSM~\cite{blukis2022persistent} & \xmark & 25.11 & 35.79 & 6.69 & 11.53 & 16.29 & 27.24 & 4.34 & 8.45 \\
FILM~\cite{min2021film} & \xmark & 25.77 & 36.15 & 10.39 & 14.17 & 24.46 & 34.75 & 9.67 & 13.13 \\
LGS-RPA~\cite{murray2022following} & \xmark & 33.01 & 41.71 & 16.65 & 24.49 & 27.80 & 38.55 & 12.92 & 20.01 \\
Prompter~\cite{inoue2022prompter} & \xmark & 47.95 & 56.98 & 23.29 & 28.42 & 41.53 & 53.69 & 18.84 & 24.20 \\
CAPEAM~\cite{kim2023context} & \xmark & 47.36 & 54.38 & 19.03 & 23.78 & 43.69 & 54.66 & 17.64 & 22.76 \\
DISCO~\cite{xu2024disco} & \xmark & 58.00 & 64.90 & 39.60 & 46.50 & 54.70 & \textbf{65.50} & 35.50 & 43.60 \\
\textbf{ESCAPE (Ours)} & \xmark & \textbf{61.24} & \textbf{68.74} & \textbf{46.89} & \textbf{53.22} & \textbf{56.04} & 64.79 & \textbf{41.82} & \textbf{48.04} \\

\thickhline
\end{tabular*}
\label{tab:sota}
\end{table}

\subsection{Comparison with the State of the Art} \label{sec:sota}
We evaluate against competitive ALFRED baselines: Seq2Seq~\cite{shridhar2020alfred}, MOCA~\cite{singh2021factorizing}, E.T.~\cite{pashevich2021episodic}, HLSM~\cite{blukis2022persistent}, ABP~\cite{kim2021agent}, FILM~\cite{min2021film}, LGS-RPA~\cite{murray2022following}, Prompter~\cite{inoue2022prompter}, CAPEAM~\cite{kim2023context}, DISCO~\cite{xu2024disco} and ThinkBot~\cite{luthinkbot}. All methods use RGB inputs and language instructions during inference. For fair comparison, we present results based on whether step-by-step instructions are used. By default, ESCAPE uses high-level goals, but can incorporate step-by-step instructions when available.

We present comprehensive comparisons with state-of-the-art methods on ALFRED benchmark in Table \ref{tab:sota}. Our ESCAPE framework demonstrates superior performance across most evaluation metrics in both instruction settings. With step-by-step instructions, ESCAPE achieves 65.09\% and 60.79\% success rates on seen and unseen test splits respectively, surpassing the previous best method ThinkBot~\cite{luthinkbot} by 2.40\% and 2.97\%. Without instructions, ESCAPE maintains strong performance with 61.24\% and 56.04\% success rates, representing improvements of 3.24\% and 1.34\% over DISCO~\cite{xu2024disco}.

Notably, ESCAPE exhibits substantial improvements in path-length-weighted metrics, achieving 52.42\% PLWSR and 58.29\% PLWGC on seen environments, and 46.82\% PLWSR and 52.57\% PLWGC on unseen environments. These significant gains validate the effectiveness of our Adaptive Execution Policy in seizing opportunistic interactions to enhance long-horizon efficiency. The consistent superior performance across almost all metrics and instruction settings demonstrates ESCAPE's robust generalization capability, indicating strong potential for real-world applications where detailed instructions may not be available.

\subsection{Ablation Study} \label{sec:expab}
We conduct ablation studies on ALFRED validation splits to evaluate each component in ESCAPE (Table \ref{tab:ablation}). Results are averaged over 3 runs with different random seeds. Detailed experimental configurations are in Appendix \ref{sec:ablation}.

\begin{table}[t]
\centering
\scriptsize
\caption{Ablation study results on ALFRED benchmark~\cite{shridhar2020alfred}. w/ GT masks denotes using ground-truth semantic segmentation masks from simulator, w/ SQ denotes using static query for interaction mask generation, and w/ Grid 80/120 denotes scene grid resolutions of 80×80 and 120×120. w/o OME, w/o MRU, and w/o AEP represent the removal of Observation-to-Memory Encoding, Memory Retrieval and Update, and Adaptive Execution Policy, respectively.}
\setlength{\tabcolsep}{4pt}
\renewcommand{\arraystretch}{1.3}
\newcommand{\thickhline}{\noalign{\hrule height 1pt}}
\newcommand{\thinhline}{\noalign{\hrule height 0.5pt}}
\begin{tabular*}{\textwidth}{@{\extracolsep{\fill}}
    p{1.8cm}|| 
    >{\centering\arraybackslash}p{0.9cm}
    >{\centering\arraybackslash}p{0.9cm}
    >{\centering\arraybackslash}p{1cm}
    >{\centering\arraybackslash}p{1cm}|| 
    >{\centering\arraybackslash}p{0.9cm}
    >{\centering\arraybackslash}p{0.9cm}
    >{\centering\arraybackslash}p{1cm}
    >{\centering\arraybackslash}p{1cm}}
\thickhline
& \multicolumn{4}{c||}{\textbf{Valid Seen}} & \multicolumn{4}{c}{\textbf{Valid Unseen}} \\
& SR & GC & PLWSR & PLWGC & SR & GC & PLWSR & PLWGC \\
\thinhline
ESCAPE & 62.32 & 66.57 & 44.24 & 49.39 & 61.27 & 69.01 & 38.06 & 42.63\\
\thinhline
w/ GT masks & 62.56 & 67.52 & 44.66 & 49.93 & 63.82 & 72.31 & 39.69 & 44.30\\
w/ SQ & 60.98 & 65.15 & 42.38 & 47.04 & 58.83 & 66.18 & 36.18 & 40.37\\
w/ Grid 80 & 52.68 & 60.08 & 40.62 & 46.22 & 52.25 & 60.94 & 30.40 & 34.47\\
w/ Grid 120 & 53.54 & 59.60 & 38.98 & 44.65 & 52.86 & 62.40 & 33.94 & 39.18\\
\thinhline
w/o OME & 53.54 & 62.30 & 38.48 & 48.03 & 52.74 & 65.47 & 35.70 & 40.51\\
w/o MRU & 42.80 & 53.53 & 19.56 & 30.51 & 27.16 & 40.42 & 9.51 & 18.43\\
w/o AEP & 57.59 & 62.74 & 38.44 & 43.63 & 57.13 & 63.39 & 34.57 & 38.07\\
\thickhline
\end{tabular*}
\label{tab:ablation}
\end{table}

\begin{table}[t]
\centering
\scriptsize
\caption{Additional quantitative metrics. mIOU scores for map and image semantic segmentation are presented, while efficiency factors (EF) are compared between ESCAPE and w/o Adaptive Execution Policy (AEP).}
\setlength{\tabcolsep}{4pt}
\renewcommand{\arraystretch}{1.3}
\newcommand{\thickhline}{\noalign{\hrule height 1pt}}
\newcommand{\thinhline}{\noalign{\hrule height 0.5pt}}
\begin{tabular*}{\textwidth}{@{\extracolsep{\fill}}
    p{1.5cm}
    >{\centering\arraybackslash}p{0.9cm}
    >{\centering\arraybackslash}p{1.0cm} ||
    p{1.5cm}
    >{\centering\arraybackslash}p{1.1cm}
    >{\centering\arraybackslash}p{1.1cm}
    >{\centering\arraybackslash}p{1.4cm}
    >{\centering\arraybackslash}p{1.4cm}}
\thickhline
\textbf{mIOU} & Seen & Unseen & \textbf{EF} & Seen SR & Seen GC & Unseen SR & Unseen GC\\
\thinhline
Map Seg. & 0.758 & 0.654 & ESCAPE & 0.710 & 0.742 & 0.621 & 0.618 \\
Image Seg. & 0.869 & 0.779 & w/o AEP & 0.668 & 0.695 & 0.605 & 0.601 \\
\thickhline
\end{tabular*}
\label{tab:extra}
\end{table}

\begin{figure}[t]
  \centering
  \includegraphics[width=0.99\textwidth]{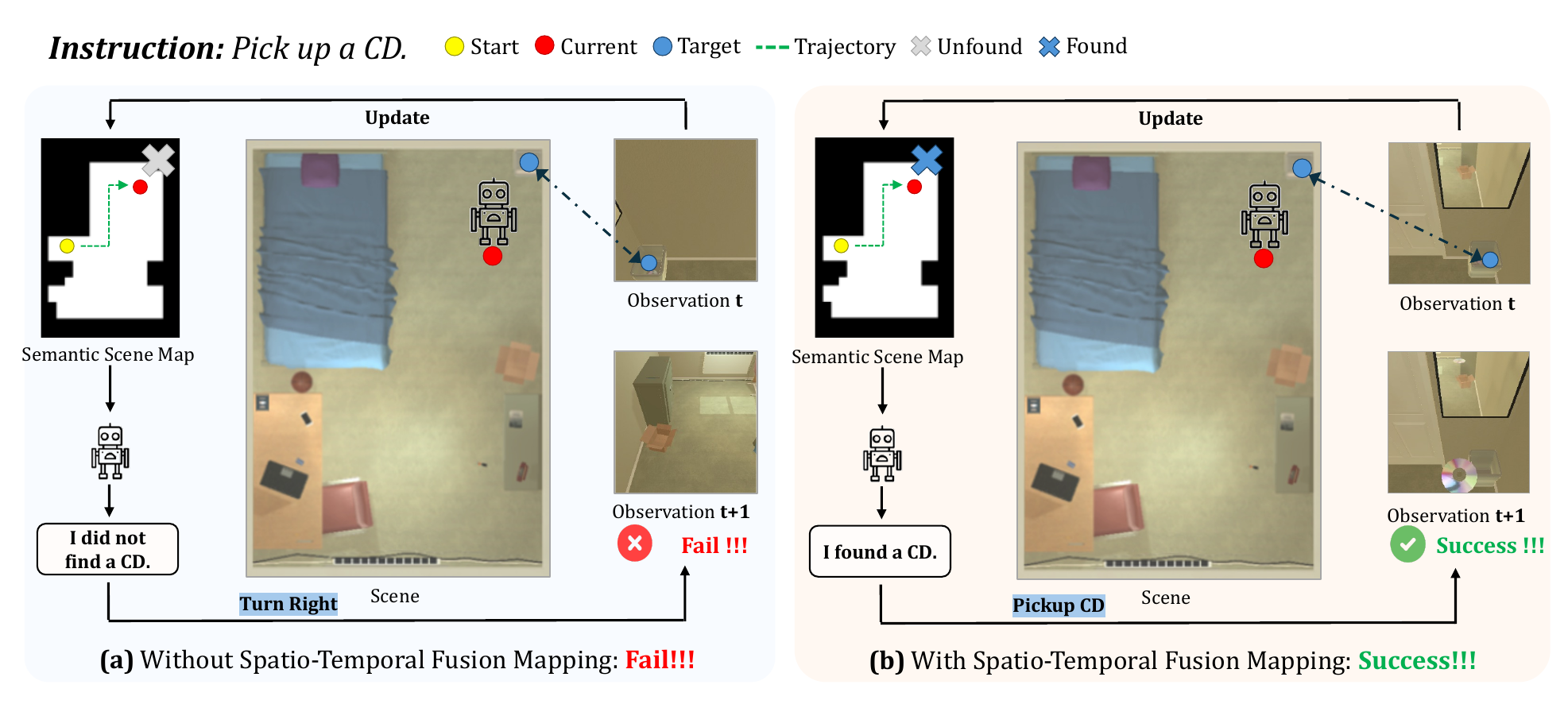}
  \caption{Qualitative analysis of our Spatio-Temporal Fusion Mapping module. \textbf{Left:} Failure due to missing spatial memory updates. \textbf{Right:} Success with the persistent 3D spatial memory.}
  \label{fig:case1}
\end{figure}

\noindent \textbf{Spatio-Temporal Fusion Mapping.} We first examine Observation-to-Memory Encoding (OME) and Memory Retrieval and Update (MRU). Removing OME causes moderate drops (8.78\% SR on seen, 8.53\% on unseen), confirming its role in fusing visual observations into 3D memory. Without MRU, severe drops occur (19.52\% SR on seen, 34.11\% on unseen), proving temporal memory retrieval and update are critical for scene understanding given the agent's limited camera view. As shown in Table \ref{tab:extra}, our map semantic segmentation achieves 0.758 mIOU on seen and 0.654 on unseen environments, indicating accurate spatial localization and robustness across novel scenarios. We also test grid resolutions of 80×80, 100×100, and 120×120. 100×100 performs best, while 80×80 and 120×120 cause significant drops (9.64\%/8.78\% SR on seen, 9.02\%/8.41\% on unseen). This confirms 100×100 best matches ALFRED's 25m×25m rooms with 25cm steps, where each grid cell equals one agent step. Both OME and MRU are essential, with MRU being more critical to avoid catastrophic forgetting.

\noindent \textbf{Memory-Driven Target Grounding.} Ablation results show the effectiveness of our Memory-Driven Target Grounding module in generating precise interaction masks. First, image semantic segmentation achieves 0.869 mIOU on seen and 0.779 on unseen environments (Table \ref{tab:extra}), showing accurate object localization. Second, comparing with GT masks shows minimal gaps (0.24\% SR on seen, 2.55\% on unseen), confirming our masks are not the bottleneck for task success. Moreover, using scene-specific 3D object features beats static queries (1.34\% SR on seen, 2.44\% on unseen), indicating that dynamically derived memory queries capture object specificity better. Furthermore, failure analysis in Appendix \ref{sec:failure} also shows our method reduces interaction failures by 12.6\% on seen and 24.3\% on unseen tasks versus DISCO~\cite{xu2024disco}.

\noindent \textbf{Adaptive Execution Policy.} Our Adaptive Execution Policy (AEP) significantly improves both task success and efficiency. Removing it causes drops: SR falls 4.73\% on seen and 4.14\% on unseen settings. To measure efficiency, we define the \textbf{Efficiency Factor (EF)}: $EF = \text{Expert Length}/\text{Agent Length} = \text{PLWSR}/\text{SR}$. Higher EF indicates more efficient task completion with shorter paths compared to expert trajectories. Table \ref{tab:extra} shows consistent EF gains across all metrics when employing AEP. These results demonstrate that dynamically shifting between global proactive planning and local opportunistic interaction yields highly efficient long-horizon execution.

\subsection{Qualitative Results}
\begin{figure}[t]
  \centering
  \includegraphics[width=0.99\textwidth]{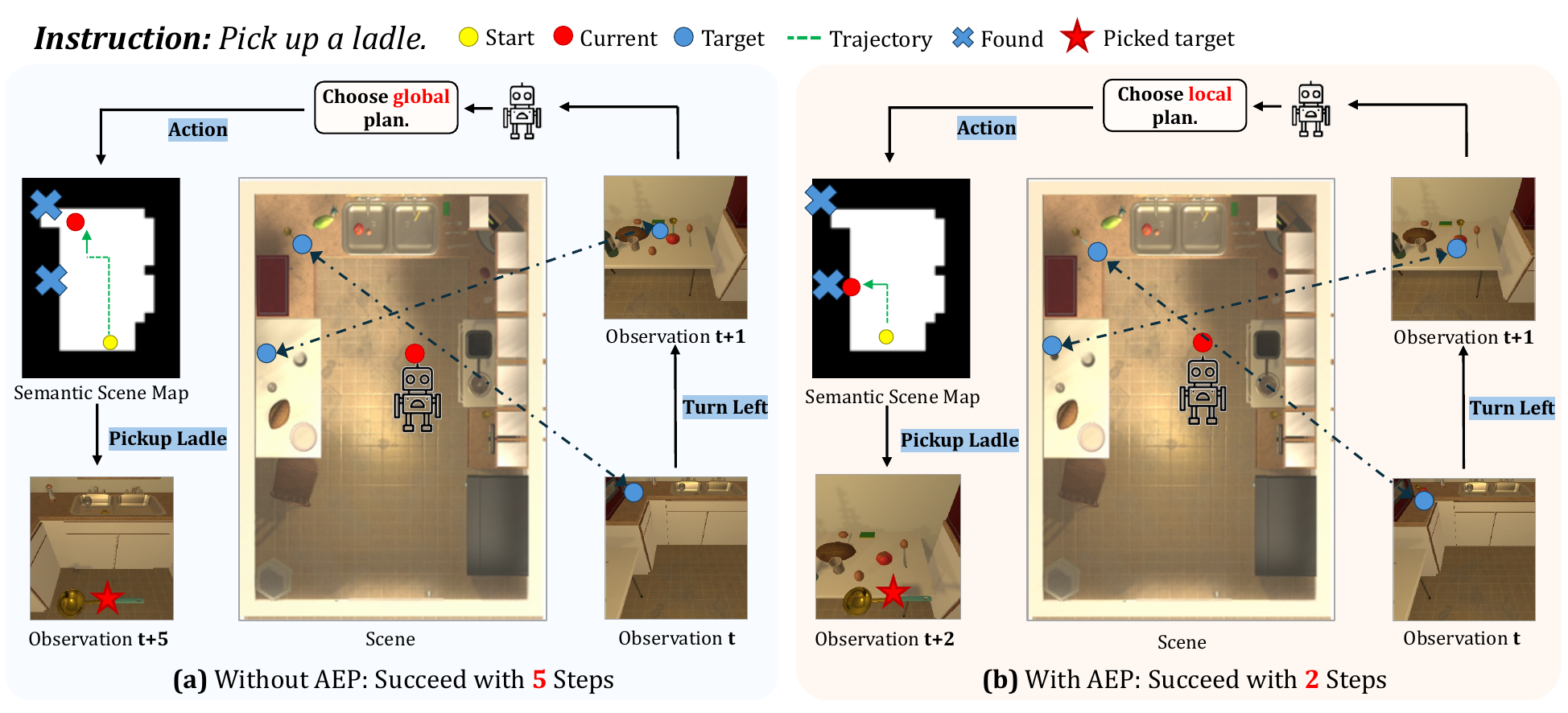}
  \caption{Qualitative comparison of the Adaptive Execution Policy (AEP) effectiveness. \textbf{Left:} Pickup task completed in 5 steps without AEP. \textbf{Right:} The same task accomplished in 2 steps with our proposed AEP.}
  \label{fig:case2}
\end{figure}

The visualization results in Fig. \ref{fig:case1} illustrates the efficacy of our Spatio-Temporal Fusion Mapping module. In scenario (a), prior approaches fail to detect the CD within the bin, demonstrating their limitations with contained objects. In contrast, scenario (b) demonstrates how our memory-centric agent successfully localizes and retrieves the CD from the bin, highlighting the module's capability in handling complex spatial relationships and containment scenarios.

Fig. \ref{fig:case2} shows the effectiveness of our Adaptive Execution Policy. Without AEP, agents strictly follow fixed long-term plans, blindly bypassing closer optimization opportunities (scenario a). In contrast, our reactive local monitor continuously screens for immediate manipulation targets and preemptively interrupts long-term routes when viable interactions are detected, greatly improving ESCAPE's efficiency (\textbf{5 steps versus 2 steps}). This behavioral improvement perfectly corroborates the substantial gains in PLW metrics shown in Table \ref{tab:sota}.

\section{Conclusion}
We present ESCAPE, a memory-centric framework for long-horizon mobile manipulation that addresses the fundamental challenges of long-horizon efficiency and dynamic execution. By coupling a persistent Episodic Spatial Memory with an Adaptive Execution Policy, ESCAPE mitigates catastrophic forgetting, spatial inconsistency, and rigid execution constraints. Extensive experiments demonstrate that ESCAPE achieves state-of-the-art performance in both seen and unseen environments on ALFRED benchmark, with significant gains in path-weighted efficiency metrics. Our ablation studies and thorough visualizations validate the contribution of each module.

\bibliographystyle{splncs04}
\bibliography{main}

% ---- Appendix ----
\clearpage
\appendix
\renewcommand{\theHsection}{A\arabic{section}}

\section{Introduction of ALFRED Benchmark} \label{sec:benchmark}
ALFRED~\cite{shridhar2020alfred} is a large scale dataset which includes 25,726 language directives for embodied mobile manipulation task. Each language directive includes a high-level goal together with low-level instructions.

Given either the goal or instructions in natural language along with egocentric vision, the agent's objective is to produce a series of actions along with object masks which guide interactions with relevant objects in order to complete the task. The agent must fulfill all the conditions required for task completion; even if one condition is not met, the task is deemed unsuccessful.

\subsection{Task Types}
All the task in ALFRED can be categorized into 7 types:

\begin{enumerate}
\item  Look \& Examine. Examine an object under the light (e.g. examine an apple under the lamp).

\item  Pick \& Place. Pick an object and place it in a receptacle (e.g. pick a bowl and place it on the counter).

\item  Pick \& Place Two. Place two object instances in the same receptacle (e.g. throw two apples into the garbage bin).

\item  Stack. Place an object in a movable container then place the movable container in a receptacle (e.g. place a fork in a plate then put the plate on the counter).

\item  Heat \& Place. Place a heated object in a receptacle (e.g. place a heated egg on the dining table).

\item  Cool \& Place. Place a cooled object in a receptacle (e.g. place a chilled potato on the dinning table).

\item  Clean \& Place. Place a cleaned object in a receptacle (e.g. place a cleaned towel in the sinkbasin).

\end{enumerate}

\subsection{Sub-tasks}
All tasks mentioned above can be divided into eight sub-tasks: (1) GotoLocation; (2) PickUp; (3) Put; (4) Slice; (5) Toggle; (6) Heat; (7) Cool; (8) Clean. Each sub-task is a pair of action and target object (e.g. (PickUp, Apple)).

\subsection{Action Space}
The action space of the agent consists of 5 navigation actions (MoveAhead, RotateRight, RotateLeft, LookUp, LookDown), 7 interactive actions (PickUp, Put, Open, Close, ToggleOn, ToggleOff, Slice) and a STOP action indicating the task is completed. Every interactive action necessitates an object mask to identify the specific object for manipulation.

\subsection{Evaluation Metrics}
The evaluation process relies on three key metrics: success rate (SR), goal-condition success rate (GC), and path-length weighted scores (PLWSR and PLWGC). The primary metric, SR, represents the proportion of tasks successfully completed, reflecting the agent’s overall task-solving capability. GC, on the other hand, measures the percentage of goal conditions that have been met, providing insight into the agent’s ability to achieve partial task success. Lastly, path-length weighted (PLW) scores adjust SR and GC based on the number of actions taken by the agent, assessing its efficiency in task completion.

\subsection{An Example Task} \label{sec:alfred_example}
\begin{figure}[t]
  \centering
  \includegraphics[width=0.99\textwidth]{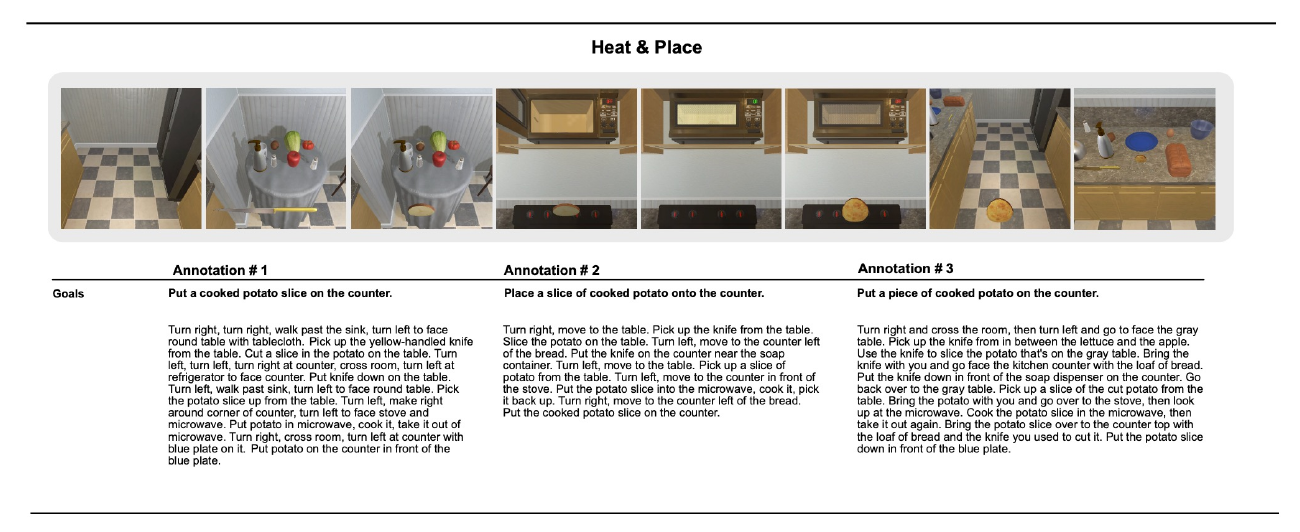}
  \caption{Illustration of an example "Heat and Place" task from the ALFRED~\cite{shridhar2020alfred} benchmark, showing the agent's sequential interactions with objects in a household environment.}
  \label{fig:alfred}
\end{figure}

As illustrated in Fig. \ref{fig:alfred}, this ALFRED example task demonstrates how the dataset provides both high-level directives and step-by-step descriptions for each task. While conventional approaches rely on detailed instructions, our ESCAPE method efficiently completes tasks using only high-level directives, eliminating the need for granular guidance while maintaining high accuracy and performance.

\section{Applying ESCAPE in ALFRED} \label{sec:application}
In this section, we describe the comprehensive process of applying ESCAPE to ALFRED.

\subsection{Language Process}
For a fair comparison, we adopt the language processing approach from~\cite{min2021film}, which utilizes fine-tuned BERT models to convert natural language instructions into ALFRED's internal parameters. While modern large language models offer potential advantages, our experimental design guarantees an equitable comparison with existing baselines.

\begin{table}[t]
\centering
\small
\renewcommand{\arraystretch}{1.1}
\newcommand{\thickhline}{\noalign{\hrule height 1pt}}
\newcommand{\thinhline}{\noalign{\hrule height 0.5pt}}
\caption{Templates on generating subgoals per task type.}
\label{tab:templates}
\begin{tabular}{ll}
\thickhline
\multicolumn{2}{l}{\textbf{(1) Look \& Examine}} \\
PickUp & Object \\
Toggle & Lamp \\
\thinhline
\multicolumn{2}{l}{\textbf{(2) Pick \& Place}} \\
PickUp & Object \\
Put & Receptacle \\
\thinhline
\multicolumn{2}{l}{\textbf{(3) Place Two}} \\
PickUp & Object \\
GotoLocation & Object \\
Put & Receptacle \\
PickUp & Object \\
Put & Receptacle \\
\thinhline
\multicolumn{2}{l}{\textbf{(4) Stack}} \\
PickUp & Object \\
Put & Movable Receptacle \\
PickUp & Movable Receptacle \\
Put & Receptacle \\
\thinhline
\multicolumn{2}{l}{\textbf{(5) Heat \& Place}} \\
PickUp & Object \\
Heat & Microwave \\
Put & Receptacle \\
\thinhline
\multicolumn{2}{l}{\textbf{(6) Cool \& Place}} \\
PickUp & Object \\
Cool & Fridge \\
Put & Receptacle \\
\thinhline
\multicolumn{2}{l}{\textbf{(7) Clean \& Place}} \\
PickUp & Object \\
Clean & Sink Basin \\
Put & Receptacle \\
\thickhline
\end{tabular}
\end{table}

First, the BERT model identifies the task type based on the language instruction, then it predicts the following four PDDL parameters: Object Target (the object to be manipulated. e.g. apple), Receptacle (where the object will be finally placed. e.g. table), Movable Receptacle (the intermediate container in stack tasks. e.g. pot), Slice (whether the object needs to be sliced.).

Building on the estimated internal parameters derived from the language module, we leverage ALFRED’s structured task framework to construct sub-tasks using predefined templates. The standard templates corresponding to each task type are provided in Table \ref{tab:templates}. Each sub-tasks consists of a verb-noun pair, which is then processed within the proposed ESCAPE framework for execution.

\subsection{Agent Setup}
The agent's 4-DoF pose is recorded, consisting of a 2-DoF position, a 1-DoF rotation, and a 1-DoF camera tilt angle. At the beginning of each task, the camera is positioned  at a 45-degree downward tilt relative to the horizon. As navigation progresses, the agent's pose is incrementally updated based on the accumulation of discrete actions.

At the beginning, the agent scans its panoramic surroundings by performing four consecutive 90-degree rotations. This process allows the agent to build a comprehensive representation of the scene, facilitating the navigation. Once this initial perception phase is complete, the agent proceeds to execute the planned sub-tasks in sequence.

\subsection{Episodic Spatial Memory}
The perception system begins with an egocentric $300 \times 300$ RGB frame, captured by a camera with a 60-degree field of view. To construct a comprehensive spatial memory, we leverage the \textit{Spatio-Temporal Fusion Mapping} module to generate a semantic scene map and the \textit{Memory-Driven Target Grounding} module to extract interaction masks. Both map semantic segmentation and image semantic segmentation utilize 87 categories. The Spatio-Temporal Fusion Mapping module utilizes Observation-to-Memory Encoding (OME) to fuse current observations with the existing spatial memory, and Memory Retrieval and Update (MRU) to retain and aggregate information from previous memory. The Memory-Driven Target Grounding module produces precise target masks for object interactions by computing pixel-wise similarity between 3D object-specific features derived from the memory and 2D image data. Details of the perception system are in Section \ref{sec:mapping} and \ref{sec:grounding} of the main text.

\subsection{Adaptive Execution Policy}
We employ an \textit{Adaptive Execution Policy} (AEP) when executing sub-tasks, which consists of a proactive global planner and a reactive local monitor. Initially, the agent randomly explores navigable areas until the target object is detected. Upon detection, the proactive global planner generates a long-horizon navigation trajectory to approach the target based on the semantic scene map generated from the spatial memory. Meanwhile, the reactive local monitor continuously screens the environment for immediate interaction opportunities. If a closer or more accessible target appears within range, it preemptively interrupts the long-term plan and proceeds with immediate interaction, significantly enhancing long-horizon execution efficiency in mobile manipulation. Details of the policy are in Section \ref{sec:policy} of the main text.

\section{Ablation Setting} \label{sec:ablation}
\textbf{ESCAPE.} Our complete model incorporates OME for spatial-visual fusion, MRU for temporal aggregation, AEP for reactive interaction, 100×100 grid resolution, and dynamically generated object queries from memory features for interaction mask prediction, without using ground-truth semantic segmentation masks from the simulator.

\noindent \textbf{w/ GT Masks.} This configuration uses ground-truth semantic segmentation masks provided by the simulator for object interaction, while maintaining all other components identical to ESCAPE (OME, MRU, AEP, 100×100 grid).

\noindent \textbf{w/ SQ.} We replace dynamically computed mask queries (generated real-time from memory features) with fixed static queries (87 object classes with 256-dimensional vectors per class that remain unchanged after training), while keeping all other components unchanged.

\noindent \textbf{w/ Grid 80/120.} We test alternative grid resolutions of 80×80 and 120×120 instead of the standard 100×100 configuration, while maintaining all other model components (OME, MRU, AEP, and dynamic queries) identical to ESCAPE.

\noindent \textbf{w/o OME.} At time $t$, we directly use the memory feature from time $t-1$ to predict semantic scene map and interaction mask, without OME updates for spatial-visual fusion, while keeping MRU, AEP, and other components active.

\noindent \textbf{w/o MRU.} At each timestep, we update memory features based solely on the current observation without incorporating historical memory features for temporal aggregation, while keeping OME, AEP, and other components unchanged.

\noindent \textbf{w/o AEP.} We eliminate the Adaptive Execution Policy mechanism that enables reactive interaction. Without this component, the agent strictly adheres to navigation routes generated by the global planner, executing a purely sequential approach to exploration and interaction.

\section{Failure Analysis} \label{sec:failure}
\begin{table}[t]
\centering
\small
\renewcommand{\arraystretch}{1.3}
\caption{Failure analysis.}
\newcommand{\thickhline}{\noalign{\hrule height 1pt}}
\newcommand{\thinhline}{\noalign{\hrule height 0.5pt}}
\label{tab:failure_analysis}
\begin{tabular}{l|c|c}
\thickhline
 & \textbf{Valid Seen} & \textbf{Valid Unseen} \\
\thinhline
Success Rate & 62.32 & 61.27 \\
\thinhline
Language error & 18.29 & 19.24 \\
Object not found & 3.91 & 2.31 \\
Navigation collision & 6.46 & 11.57 \\
Interaction failure & 9.02 & 5.61 \\
\thickhline
\end{tabular}
\end{table}

We conduct a comprehensive failure analysis of ESCAPE as reported in Table \ref{tab:failure_analysis}. ESCAPE achieves 62.32\% success rate on seen environments and 61.27\% on unseen environments, demonstrating consistent performance across different scenarios. Language misunderstanding errors constitute the largest proportion of failures, accounting for approximately 18-19\% in both settings. This suggests the necessity of more robust language understanding capabilities in embodied AI systems. Object not found failures occur in 3.91\% and 2.31\% of cases respectively, likely due to target objects being located in closed receptacles or occluded areas that require active exploration with commonsense reasoning. Navigation collision and interaction failure represent other common failure modes, with navigation issues being more prominent in unseen environments (11.57\% vs 6.46\%), indicating the challenge of generalizing spatial navigation to novel scenes.

\end{document}